\pdfoutput=1

\documentclass[11pt]{article}

\usepackage[]{acl}

\usepackage{times}
\usepackage{latexsym}

\usepackage[T1]{fontenc}

\usepackage[utf8]{inputenc}

\usepackage{microtype}

\usepackage{inconsolata}

\usepackage{graphicx}
\usepackage{amsfonts}
\usepackage{array}
\usepackage{multirow}
\usepackage{comment}
\usepackage{subcaption}
\usepackage{amsmath}
\usepackage{rotating}

\newcolumntype{H}{>{\arraybackslash}m{2.25cm}}
\newcolumntype{M}{>{\arraybackslash}m{2.8cm}}
\newcolumntype{K}{>{\arraybackslash}m{2.5cm}}
\newcolumntype{G}{>{\arraybackslash}m{3.9cm}}
\newcolumntype{T}{>{\centering\arraybackslash}m{0.35cm}}
\newcolumntype{N}{>{\centering\arraybackslash}m{0.3cm}}
\newcolumntype{Z}{>{\centering\arraybackslash}m{1.5cm}}

\usepackage{nicefrac}
\usepackage{color,soul}
\usepackage{xcolor,colortbl}

\title{BoK: Introducing Bag-of-Keywords Loss \\for Interpretable Dialogue Response Generation}

\author{Suvodip Dey \and Maunendra Sankar Desarkar\\
   Indian Institute of Technology Hyderabad, India \\
   \texttt{suvodip15@gmail.com,  maunendra@cse.iith.ac.in} \\
   }

\begin{document}
\maketitle
\begin{abstract}

The standard language modeling (LM) loss by itself has been shown to be inadequate for effective dialogue modeling. As a result, various training approaches, such as auxiliary loss functions and leveraging human feedback, are being adopted to enrich open-domain dialogue systems. One such auxiliary loss function is Bag-of-Words (BoW) loss, defined as the cross-entropy loss for predicting all the words/tokens of the next utterance. In this work, we propose a novel auxiliary loss named Bag-of-Keywords (BoK) loss to capture the central thought of the response through keyword prediction and leverage it to enhance the generation of meaningful and interpretable responses in open-domain dialogue systems. BoK loss upgrades the BoW loss by predicting only the keywords or critical words/tokens of the next utterance, intending to estimate the core idea rather than the entire response. We incorporate BoK loss in both encoder-decoder (T5) and decoder-only (DialoGPT) architecture and train the models to minimize the weighted sum of BoK and LM (BoK-LM) loss. We perform our experiments on two popular open-domain dialogue datasets, DailyDialog and Persona-Chat. We show that the inclusion of BoK loss improves the dialogue generation of backbone models while also enabling post-hoc interpretability. We also study the effectiveness of BoK-LM loss as a reference-free metric and observe comparable performance to the state-of-the-art metrics on various dialogue evaluation datasets.

\end{abstract}

\section{Introduction}
Open-domain dialogue generation is a dynamic area of research, aiming to generate contextually relevant and meaningful responses given a dialogue context. As deep learning models continue to thrive in the field of natural language processing (NLP), a widely adopted strategy to solve any natural language generation (NLG) task involves pre-training and fine-tuning large language models (LLMs). The LLMs are predominantly trained with language modeling (LM) loss, which essentially corresponds to cross-entropy loss for predicting the next word or token. While LM loss remains effective in training NLG models for diverse tasks, including dialogue generation \cite{hred, tranfertransfo, dialogpt, blenderbot}, it may not be the optimal choice for training models specifically tailored for dialogue generation. It is well-established that perplexity, a measure associated with LM loss, primarily gauges fluency and weakly correlates with human dialogue evaluation \cite{convai2, usr,usl-h}. Consequently, relying solely on LM loss may not guarantee generations with desirable conversational qualities. Therefore, exploring alternative loss functions and training methods is crucial to advance the development of generative open-domain dialogue models.

\begin{figure}[t]
    \centering
    \includegraphics[scale=0.7]{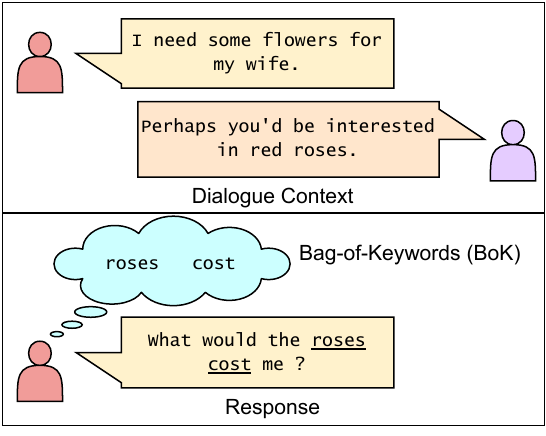}
    \caption{A motivating example for Bag-of-Keywords loss in open-domain dialogue system.}
    \label{fig:concept}
\end{figure}

In order to mitigate the exclusive dependence on LM loss in the training of open-domain dialogue models, various approaches have been explored in the existing literature. These techniques can be broadly categorized into two classes - a) auxiliary loss and b) human feedback. The first approach combines one or more auxiliary losses with LM loss to train the dialogue models. Various types of auxiliary losses have been explored in the context of open-domain dialogue learning. For instance, Bag-of-Words (BoW) loss computes the cross-entropy loss to predict words/tokens of the next utterance from the given dialogue context \cite{bow_loss, dialoflow, dialogen}. Some methodologies involve predicting the sentence-level encoding of the next utterance and determining the loss through L1/L2 norms and KL divergence \cite{vhred, dialoflow, dialogved, dialogen}. Few approaches incorporate a next-utterance classification loss \cite{tranfertransfo}, wherein the auxiliary loss is computed for a classification or ranking task to predict the true utterance from a set of candidate responses. On the other hand, the second approach is based on refining the pre-trained dialogue model through human feedback. These methods mostly follow the training principle of Reinforcement learning from human feedback (RLHF), where the model is fine-tuned to maximize the reward associated with the generated response using Reinforcement learning. RLHF has gained significant interest recently, particularly with the popularity of models like Chat-GPT \cite{instruct-gpt}. However, acquiring quality human feedback data is challenging and expensive \cite{rlhf-problem}. Furthermore, relying on automated dialogue evaluation metrics as a substitute for human feedback can pose challenges, as they may not strongly correlate with human judgments \cite{liu-etal-2016-evaluate, metric-survey}.

In this work, our objective is to propose a novel auxiliary loss for open-domain dialogue systems. Specifically, we address the limitation of BoW loss by introducing Bag-of-Keywords (BoK) loss, which is defined as the cross-entropy loss to predict the keywords of the next utterance. While training, we extract the keywords of the ground-truth response using YAKE!~\cite{yake2, yake}, an unsupervised feature-based keyword extractor. The keywords can be seen as a proxy for the core idea of the response. In a conversation, a reply can be generated in multiple ways. As a result, BoW loss can induce training data bias since it considers all the words/tokens of the ground-truth response for prediction. In contrast, BoK loss focuses on the core idea (as shown in Fig.~\ref{fig:concept}) that alleviates the problem of generalization. The main contributions of this work are summarized as follows\footnote[1]{Code is available at \href{https://github.com/SuvodipDey/BoK}{github.com/SuvodipDey/BoK}}:
\begin{itemize}
    \item We propose BoK loss, a novel auxiliary loss for open-domain dialogue systems. BoK loss can be easily incorporated into any generative model and trained using a weighted sum of BoK and LM (BoK-LM) loss. 
    \item We show that BoK loss enhances the dialogue generation of backbone models on DailyDialog and Persona-Chat datasets. We note an improvement in the specificity of the generated responses with the inclusion of BoK loss. 
     \item We perform a qualitative analysis of the generated responses and discuss how BoK loss enables post-hoc interpretability. 
    \item We study the effectiveness of BoK-LM loss as a reference-free metric. We observe that it exhibits moderate correlations with human judgments on different evaluation datasets. 
\end{itemize}

\section{Background and Related Works}

Open-domain dialogue generation is a challenging NLG task. Let $D_{<t}=\{u_1, u_2, ... u_{t-1}\}$ be a multi-turn conversation where $u_j$ represents the utterance at turn $j$. Let $C_t$ be the condition (like persona, document, etc.) other than dialogue history for generating $u_t$. The task of open-domain dialogue generation is to generate $u_t$ given $D_{<t}$ and $C_t$. Like any NLG task, it is modeled using language models and generally trained using the next word/token prediction task. The corresponding language modeling (LM) loss is defined as,
\begin{equation} \label{eqn:lm}
\mathcal{L}_{\operatorname{LM}} = - \sum_{n=1}^{T} \log p(u_{{t}_{n}}|u_{{t}_{<n}}, D_{<t}, C_t; \theta)
\end{equation}
where $u_{{t}_{n}}$ denotes the $n^{th}$ word/token of utterance $u_t$ and $\theta$ indicates the parameters of the language model. Training transformer~\cite{transformer} based large language models (LLMs) with LM loss on large dialogue corpora has shown remarkable performance in open-domain dialogue generation \cite{dialogpt, blenderbot}. However, it has been shown that perplexity ($e^{\mathcal{L}_{\operatorname{LM}}}$), a metric that is a function of LM loss, can measure fluency but shows a weak correlation with other conversational aspects \cite{convai2, usr,usl-h}. The root cause of this behavior stems from the inherent one-to-many nature of dialogue, where a given context can elicit multiple possible responses \cite{liu-etal-2016-evaluate}. Consequently, simply increasing the size of training data may not always yield improvement, as it is impractical to collect all potential response variations. To tackle this challenge, researchers employ various techniques, broadly categorized into two classes: i) incorporating one or more auxiliary losses alongside LM loss, and ii) leveraging human feedback to finetune pre-trained dialogue models. Given our focus on proposing a new auxiliary loss, we keep our related works limited to different auxiliary losses utilized for open-domain dialogue generation, described as follows.

\begin{figure*}[ht!]
\centering
\begin{subfigure}{.5\textwidth}
  \centering
  \includegraphics[scale=0.72]{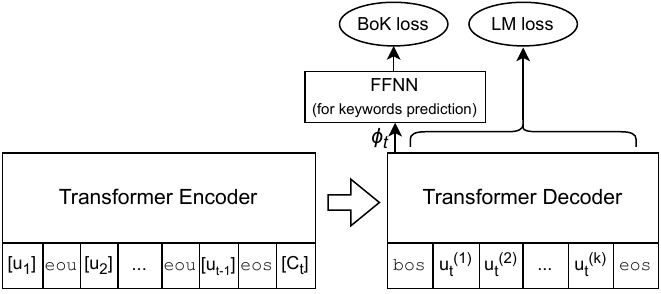}
  \caption{Encoder-decoder framework}
  \label{fig:sub1}
\end{subfigure}%
\begin{subfigure}{.5\textwidth}
  \centering
  \includegraphics[scale=0.72]{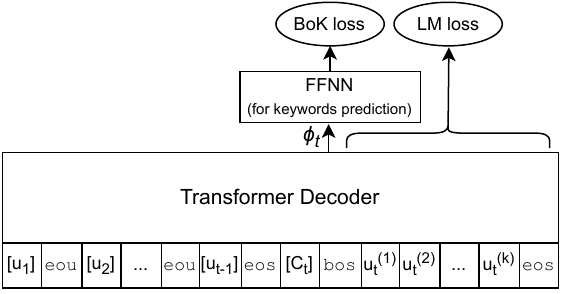}
  \caption{Decoder-only framework}
  \label{fig:sub2}
\end{subfigure}
\caption{Incorporating BoK loss in open-domain dialogue models. $[u_j]$ and $[C_j]$ represents the list of tokens after tokenizing utterance $u_j$ and condition $C_j$, respectively. $u_t^{(i)}$ denotes the $i^{th}$ token of utterance $u_t$, whereas \{$\operatorname{eos}$, $\operatorname{bos}$, $\operatorname{eou}$\} are special tokens. $\phi_{t} \in  \mathbb{R}^{d}$ is the hidden state of the final layer of $\operatorname{bos}$ token, representing the context.}
\label{fig:arc}
\end{figure*}

\begin{itemize}
    \item The first kind of auxiliary loss estimates the error in predicting the sentence-level encoding of the next utterance given the dialogue context. Authors of VHRED \cite{vhred} and DialogVED \cite{dialogved} use Kullback-Leibler (KL) divergence to compute the distance between the approximate and true posterior distribution of the next utterance. Models like DialoFlow \cite{dialoflow} and DialoGen \cite{dialogen} use the L1/L2 norm for the same purpose. Predicting the encoding of the next utterance is challenging and may lead to issues like posterior collapse while using KL divergence \cite{dialogved}.
    \item The second approach is based on the next utterance classification loss. In this method, the task is to classify the ground-truth response from a given set of candidate utterances \cite{tranfertransfo}. It is worth noting that this method requires negative samples, which are usually not included in the datasets. Hence, different kinds of negative sampling techniques are adopted to obtain them. However, finding high-quality negative samples is difficult for dialogues \cite{pone}.
    \item The third approach focuses on predicting the words/tokens of the next utterance. This loss is popularly known as Bag-of-Words (BoW) loss \cite{bow_loss}. Models like DialoFlow \cite{dialoflow} and DialoGen \cite{dialogen} utilize BoW loss to support LM loss. DialogVED \cite{dialogved}  uses BoW loss to tackle the posterior collapse that is caused due to minimizing KL divergence. As discussed earlier, a dialogue context can have many relevant responses. Hence, the task of predicting all the words/tokens can induce training data bias. In this work, we aim to address this limitation of BoW loss. 
\end{itemize}

\section{Methodology}
In this section, we describe Bag-of-Keywords loss followed by its application in open-domain dialogue systems.

\subsection{Bag-of-Keywords (BoK) loss}
As discussed, BoW loss is computed as the cross-entropy loss to predict all the tokens of the next utterance. Say the model has to generate utterance $u_t$ given dialogue context $D_{<t}$. Let $\phi_{t} \in \mathbb{R}^{d}$ be the representation of the context for generating $u_t$. Then the BoW loss ($\mathcal{L}_{\operatorname{BoW}}$) is defined as,
\begin{equation} \label{eqn:bow}
    \mathcal{L}_{\operatorname{BoW}} = -\sum_{w \in u_t}^{} \log p(w | \phi_{t})
\end{equation}
where $p(w | \phi_{t})$ is the probability of predicting the word/token $w \in u_t$ given $\phi_{t}$. Predicting all the words of a dialogue response may cause training data bias because there can be multiple ways to generate a response. Additionally, dialogue responses often contain stopwords that are necessary for sentence construction and fluency. Therefore, predicting these stopwords in BoW loss is unnecessary since LM loss already takes care of it.

One simple approach to address this limitation of BoW loss is to predict only the keywords of the response. By keywords, we mean the critical words that capture the core concept of the response. This approach can help reduce the training data bias and increase its generalizability for open-domain dialogue generation. To achieve this, we propose Bag-of-Keywords (BoK) loss, which is computed as the cross-entropy loss to predict the keywords of the next utterance. We define BoK loss ($\mathcal{L}_{\operatorname{BoK}}$) as,
\begin{equation} \label{eqn:bok}
    \mathcal{L}_{\operatorname{BoK}} = -\sum_{w \in K_t}^{} \log p(w | \phi_{t})
\end{equation}
where $K_t$ is the set of keywords (or tokens associated with the keywords) in $u_t$. Note that the annotations regarding the keywords are not available in the existing dialogue datasets. In this work, we find the keywords using YAKE! \cite{yake2, yake}, an unsupervised feature-based keyword extraction algorithm that leverages statistical features extracted directly from the text, thereby supporting texts of multiple domains and languages. However, one can adopt any strategy for keyword extraction. We chose YAKE! because it is unsupervised and has already been utilized to extract keywords from dialogue responses \cite{dial-m}. 
For example, in Fig.~\ref{fig:concept}, YAKE! extracted the keywords ``\textit{roses}'' and ``\textit{cost}'' from the response ``\textit{What would the roses cost me?}''.

\subsection{Application of BoK loss}
\label{sec:app}
BoK loss can be easily applied to any open-domain dialogue model. Currently, all state-of-the-art dialogue generation models are based on Transformer \cite{transformer}. These models can be broadly classified into two architectures - i) encoder-decoder and ii) decoder-only. Incorporating BoK loss into both these architectures is described as follows.

\begin{itemize}
    \item \textbf{Encoder-Decoder Architecture}: Fig. \ref{fig:sub1} shows the method of applying BoK loss in encoder-decoder architecture. The encoder takes the concatenation of the past utterances ($D_{<t}$) along with the condition $C_t$ as input. Note that $C_t$ may be present or absent based on the task or dataset. In the decoder, we add an extra component for computing the BoK loss. Let $\phi_{t} \in  \mathbb{R}^{d}$ be the hidden state representation of the final layer corresponding to the $\operatorname{bos}$ token, representing the context. Then, the BoK loss is computed as follows:
    \begin{equation} \label{eqn:4}
        \alpha_{t} = \mathrm{softmax}(\mathrm{FFNN}(\phi_{t})) \in \mathbb{R}^{|V|}
    \end{equation}
    \begin{equation} \label{eqn:5}
        \mathcal{L}_{\operatorname{BoK}} = -\sum_{w \in K_t}^{} \log p(w | \phi_{t}) = -\sum_{w \in K_t}^{}\log \alpha_{{t}_w}
    \end{equation}
    where $\mathrm{FFNN}$ denotes a single layer feed-forward neural network, and $|V|$ is the vocabulary size of the decoder tokens.

    \item  \textbf{Decoder-only Architecture}: Fig. \ref{fig:sub2} shows the process of incorporating BoK loss in decoder-only architecture. The BoK loss computation follows the same equations (Eqn. \ref{eqn:4} and \ref{eqn:5}) as encoder-decoder architecture.  
\end{itemize}

The training objective for both architectures is to minimize the weighted sum of BoK and LM loss. We term this loss as BoK-LM loss ($\mathcal{L}_{\operatorname{BoK-LM}}$).
\begin{equation}  \label{eqn:boklm}
    \mathcal{L}_{\operatorname{BoK-LM}} = \mathcal{L}_{\operatorname{LM}} + \lambda \mathcal{L}_{\operatorname{BoK}}
\end{equation}
where $\lambda \in \mathbb{R}$ is a hyper-parameter to set the weight of the BoK loss. Note that both the loss components depend on the context vector $\phi_t$. Hence, the BoK-LM loss helps to learn $\phi_t$ such that it can capture the core idea of the response and align the generation towards a meaningful response.

\begin{table}[t]
\centering
\begin{scriptsize}

\begin{tabular}{m{1.15cm}|c|r|r|r|r|r}
\hline \textbf{Dataset} & \textbf{Type} & \textbf{\#Dialog} & \textbf{\#Turns} & \textbf{T\textsubscript{max}} & \textbf{T\textsubscript{min}} & \textbf{T\textsubscript{avg}} \\ \hline
\multirow{3}{*}{DailyDialog} &
Train & 11118 & 87170 & 35 & 2 & 7.84\\
&
Dev & 1000 & 8069 & 31 & 2 & 8.07\\
&
Test & 1000 & 7740 & 26 & 2 & 7.74\\
\hline
\multirow{3}{*}{Persona-Chat} &
Train & 8939 & 131438 & 50 & 12 & 14.70 \\
&
Dev & 1000 & 15602 & 26 & 14 & 15.60 \\
&
Test & 968 & 15024 & 34 & 14 & 15.52 \\ \hline

\end{tabular}
\caption{Basic statistics of DailyDialog and Persona-Chat dataset. T\textsubscript{max}, T\textsubscript{min}, and T\textsubscript{avg} indicate maximum, minimum, and average dialogue turns.}
\label{tbl:stat}
\end{scriptsize}
\end{table}

\section{Experimental Set up}

\subsection{Datasets}
We conduct our experiments on two datasets: DailyDialog \cite{dailydialog} and Persona-Chat \cite{persona-chat}. 
DailyDialog is a popular chit-chat dataset in which the task is to generate responses conditioned only on the dialogue history. 
\begin{table*}[t]
\begin{scriptsize}
\centering
\begin{tabular}{p{1.2cm}|rrrr|rr|r|rr|r|TTTr}
\hline 

\multirow{2}{*}{Model}&\multicolumn{10}{c|}{Referenced Metrics}&\multicolumn{4}{c}{Reference-Free Metric} \\ \cline{2-15}
 & \textbf{Bleu-1} & \textbf{Bleu-2} & \textbf{Bleu-3} & \textbf{Bleu-4} & \textbf{Nist-2} & \textbf{Nist-4} & \textbf{Meteor} & \textbf{Div-1} & \textbf{Div-2} & \textbf{Entropy} & \textbf{U} & \textbf{S} & \textbf{L\textsubscript{S}} & \textbf{USL\textsubscript{S}-H}\\ \hline
DialoFlow & 48.75 & 26.73 & 16.35 & 10.70 & 3.76 & 3.97 & 16.44 & 0.039 & 0.216 & \underline{9.98} & 0.96 & 0.88 & 0.21 & 0.6777 \\
DialoGen &  49.13 & 27.25 & 16.88 & 11.07 & 3.76 & 3.98 & 16.40 & 0.043 & 0.223 & 9.88 & 0.83 & 0.90 & 0.32 & 0.6685 \\ 
DialogVED & 50.50 & 28.95 & 18.38 & 12.29 & 3.94 & 4.18 & \underline{16.90} & 0.037 & 0.204 & 9.82 & 0.86 & 0.88 & 0.30 & 0.6642\\
\hline
T5 &  51.56 & 29.22 & 18.29 & 12.05 & 3.99 & 4.23 & 16.27 & 0.044 & 0.219 & 9.62 & 0.97 & 0.89 & 0.18 & 0.6718\\ 
T5\textsubscript{BoW}  &  \textbf{51.75} & \underline{29.70} & 18.89 & 12.75 & \underline{4.05} & 4.32 & 16.64 & 0.045 & 0.230 & 9.79 & 0.97 & 0.89 & 0.19 & 0.6791\\ 
T5\textsubscript{BoK} &  \underline{51.74} & \textbf{29.74} & \underline{19.19} & \underline{13.24} & \textbf{4.09} & \textbf{4.37} & 16.62 & \underline{0.045} & \underline{0.233} & 9.84 & 0.97 & 0.90 & 0.20 & \underline{0.6793} \\ 
\hline
DialoGPT & 49.30 & 27.63 & 17.37 & 11.68 & 3.78 & 4.01 & 16.67 & 0.037 & 0.193 & 9.66 & 0.97 & 0.89 & 0.19 & 0.6731 \\
DialoGPT\textsubscript{BoW}  &  49.60 & 27.85 & 17.60 & 11.82 & 3.80 & 4.04 & 16.83 & 0.037 & 0.190 & 9.60 & 0.97 & 0.89 & 0.20 & 0.6759 \\
DialoGPT\textsubscript{BoK}  &  49.16 & 29.10 & \textbf{20.00} & \textbf{14.92} & 4.01 & \underline{4.35} & \textbf{17.72} & \textbf{0.048} & \textbf{0.257} & \textbf{10.19} & 0.97 & 0.89 & 0.31 & \textbf{0.7064} \\
\hline

\end{tabular}
\caption{Comparison of dialogue generation performance on DailyDialog test data with automated metrics. The highest and second-highest scores are written in bold and underlined respectively.}
\label{table:result1}
\end{scriptsize}
\end{table*} 
On the other hand, Persona-Chat is a knowledge-grounded dataset where a response needs to be generated based on both dialogue history and a persona profile that defines the speaker. Table \ref{tbl:stat} displays the basic statistics of the two datasets.





\subsection{Implementation Details}
\label{sec:det}
We choose T5 \cite{T5} and DialoGPT \cite{dialogpt} as our encoder-decoder and decoder-only architecture, respectively. We perform our experiments with T5-large\footnote[1]{\href{https://huggingface.co/google-t5/t5-large}{huggingface.co/google-t5/t5-large}} ($\approx$770M parameters) and DialoGPT-large \footnote[2]{\href{https://huggingface.co/microsoft/DialoGPT-large}{huggingface.co/microsoft/DialoGPT-large}} ($\approx$774M parameters) for both DailyDialog and Persona-Chat datasets. All the implementations are done using PyTorch and Huggingface \cite{huggingface} libraries in Python 3.10, and executed on a Nvidia A100 with 40GB memory. We use AdamW optimizer with a learning rate of 5e-5, batch size of 16, maximum training epochs of 20, and early stopping to train the models. We use beam search with a beam width of 5, maximum sequence length of 40, minimum sequence length of 11, and length penalty of 0.1 to generate responses for all the models. The rest of the details are provided in Appendix~\ref{sec:app1}.




\subsection{Baselines}
We refer to T5 and DialoGPT trained with BoK-LM loss as T5\textsubscript{BoK} and DialoGPT\textsubscript{BoK}, respectively. We compare them with vanilla T5 and DialoGPT models, trained only with LM loss. To measure the improvement over BoW loss, we also train T5 and DialoGPT with a weighted sum of BoW and LM loss (like Eqn. \ref{eqn:boklm}), denoted as T5\textsubscript{BoW} and DialoGPT\textsubscript{BoW} respectively. We also have some dataset-specific baselines. For DailyDialog, we use DialoFlow~\cite{dialoflow}, DialogVED~\cite{dialogved}, and DialoGen~\cite{dialogen}. All these three baselines use BoW loss and sentence-level next utterance prediction loss. For Persona-Chat, we use TransferTransfo \cite{tranfertransfo} and DialogVED. TransferTransfo utilizes the next utterance classification as the auxiliary loss.

\begin{table}[t]
\begin{scriptsize}
\centering
\begin{tabular}{l|rrrr|r}
\hline \textbf{Model} & \textbf{U} & \textbf{S} & \textbf{L\textsubscript{S}} & \textbf{USL\textsubscript{S}-H} & \textbf{Dial-M} \\ 
\hline
TransferTransfo & 0.75 & 0.63 & 0.44 & 0.5502 & 1.7730 \\
DialogVED  & 0.74 & 0.84 & 0.38 & \textbf{0.6348} & 1.7499 \\
\hline
T5 & 0.71 & 0.73 & 0.39 & 0.5756 & 0.9288 \\
T5\textsubscript{BoW}  & 0.72 & 0.75 & 0.40 & 0.5867 & \underline{0.8781} \\
T5\textsubscript{BoK} & 0.72 & 0.76 & 0.41 & \underline{0.5947} & \textbf{0.8556}  \\
\hline
DialoGPT & 0.76 & 0.72 & 0.36 & 0.5788 & 1.0312 \\
DialoGPT\textsubscript{BoW}  & 0.77 & 0.71 & 0.40 & 0.5868 & 1.0013 \\
DialoGPT\textsubscript{BoK} & 0.77 & 0.72 & 0.42 & 0.5923 & 1.0004 \\

\hline
\end{tabular}
\caption{Comparison of dialogue generation performance on Persona-Chat test data.}
\label{tbl:persona}
\end{scriptsize}
\end{table}

\section{Results and Analysis}
\subsection{DailyDialog Dataset}
Table~\ref{table:result1} compares the performance of various models on DailyDialog test data. We use BLEU~\cite{bleu}, NIST~\cite{nist}, METEOR~\cite{meteor}, Diversity~\cite{diversity}, and Entropy~\cite{entropy} for referenced evaluation, and USL-H~\cite{usl-h} for reference-free evaluation. As word-overlapping based metrics are not reliable with only one reference, we conduct the referenced evaluation using multi-reference DailyDialog~\cite{multi-ref} that contains four additional references along with the original response. For BoK loss, we set the maximum number of keyword tokens $|K_t|=8$ (refer Eqn. \ref{eqn:bok}). For BoK-LM loss in Eqn. \ref{eqn:boklm}, we set $\lambda$ to 0.1 and 0.3 for T5 and DialoGPT architecture, respectively. The effect of varying $\lambda$ and $|K_t|$ is studied in the ablation study. The key observations from Table~\ref{table:result1} are discussed below.

\begin{table*}[t]
\centering
\begin{scriptsize}

\begin{tabular}{l|l||r|r|r||r|r|r||r|r|r||r|r|r||r|r|r}
\hline 
\multirow{2}{*}{\textbf{Comparisons}} & \multirow{2}{*}{{\textbf{Dataset}}} & \multicolumn{3}{c|}{\textbf{Coherence}} & \multicolumn{3}{c|}{\textbf{Engagingness}} & \multicolumn{3}{c|}{\textbf{Informativeness}} & \multicolumn{3}{c|}{\textbf{Interactiveness}} & \multicolumn{3}{c}{\textbf{Overall}}\\ \cline{3-17}
& & \textbf{W} & \textbf{L} & \textbf{T} & \textbf{W} & \textbf{L} & \textbf{T}  & \textbf{W} & \textbf{L} & \textbf{T} & \textbf{W} & \textbf{L} & \textbf{T} & \textbf{W} & \textbf{L} & \textbf{T} \\ \hline

\multirow{2}{*}{T5\textsubscript{BoK} vs. T5\textsubscript{BoW} } & 
DailyDialog & 24 & 18 & 58 & 30 & 26 & 44 & 20 & 14 & 66 & 26 & 18 & 56 & 32 & 26 & 42 \\ \cline{2-17}
& 
Persona-Chat & 26 & 18 & 56 & 24 & 20 & 56 & 24 & 18 & 58 & 20 & 18 & 62 & 28 & 24 & 48 \\
\hline
\multirow{2}{*}{DialoGPT\textsubscript{BoK} vs. DialoGPT\textsubscript{BoW} } & 
DailyDialog & 42 & 34 & 24 & 30 & 30 & 40 & 44 & 26 & 30 & 34 & 30 & 36 & 46 & 34 & 20 \\ \cline{2-17}
& 
Persona-Chat & 28 & 18 & 54 & 14 & 20 & 66 & 24 & 18 & 58 & 14 & 16 & 70 & 28 & 22 & 50 \\
\hline

\end{tabular}
\caption{Human evaluation for comparing the impact of BoK and BoW loss on the performance of the backbone models. ``W'', ``L'', and ``T'' denote the percentage of win, loss, and tie, respectively.}
\label{tbl:human_eval}
\end{scriptsize}
\end{table*}

\textbf{Referenced Evaluation:} Firstly, we observe that the inclusion of BoW loss enhances the performance of both vanilla T5 and DialoGPT across all metrics. BoW loss is optimized to predict all the words/tokens of the next utterance, thereby improving the unigram match i.e. Bleu-1 score. Our findings corroborate this observation, demonstrating that T5\textsubscript{BoW} and DialoGPT\textsubscript{BoW} attain higher Bleu-1 scores compared to their other counterparts. Secondly, we note that both T5\textsubscript{BoK} and DialoGPT\textsubscript{BoK} perform better than their BoW counterpart in most of the cases. Furthermore, they also outperform the three baselines (DialoFlow, DialoGen, and DialogVED) that rely on BoW loss. This indicates that BoK loss effectively improves the generalizability of BoW loss, making it more efficient.

\textbf{Reference-free Evaluation:} We use USL-H as our reference-free metric, which is a combination of three sub-metrics - Understandability (U), Sensibility (S), and Likability (L). We specifically make use of the USL\textsubscript{S}-H variant, where the likability of a response is captured through its specificity. USL\textsubscript{S}-H estimates understandability, sensibility, and specificity using valid prediction, next utterance prediction, and MLM task, respectively~\cite{usl-h}. Similar to the results of the referenced evaluation, T5\textsubscript{BoK} and DialoGPT\textsubscript{BoK} achieve better USL\textsubscript{S}-H scores than their other counterparts. Moreover, we note that for T5\textsubscript{BoK} and DialoGPT\textsubscript{BoK}, USL\textsubscript{S}-H improves because of the likability or specificity aspect. We also observe this behavior in Table~\ref{tbl:persona}, which indicates that incorporating BoK loss enhances the specificity of the generated responses.

\subsection{Persona-Chat Dataset}
The results of the Persona-Chat test data are presented in Table~\ref{tbl:persona}. Unlike DailyDialog, Persona-Chat does not have any multi-referenced test data. Therefore, we use only reference-free metrics to ensure a fair evaluation. In addition to USL\textsubscript{S}-H, we also evaluate using Dial-M~\cite{dial-m}, a masking-based reference-free metric that is effective in evaluating knowledge-grounded dialogues. It is worth mentioning that in Dial-M, a lower score is indicative of better performance as it is based on cross-entropy loss. In Table~\ref{tbl:persona}, we again observe that T5\textsubscript{BoK} and DialoGPT\textsubscript{BoK} attain better USL\textsubscript{S}-H and Dial-M scores than their other counterparts. Furthermore, we observe that DialogVED outperforms all the models on USL\textsubscript{S}-H. This is because it does not use persona profiles explicitly and relies on specially trained latent variables (on next utterance prediction) for persona-grounded response generation. Furthermore, USL\textsubscript{S}-H only considers dialogue history as context and ignores persona. As a result, DialogVED performs better in understandability and sensibility, which are estimated using valid and next utterance prediction tasks, respectively. However, it falls short in specificity and Dial-M as it does not use persona. 

\begin{table*}[t]
\begin{scriptsize}
\centering
\begin{tabular}{p{0.5cm}|N|rrrr|rr|r|rr|r|NNNr}
\hline \textbf{Model} & \textbf{$\lambda$} & \textbf{Bleu-1} & \textbf{Bleu-2} & \textbf{Bleu-3} & \textbf{Bleu-4} & \textbf{Nist-2} & \textbf{Nist-4} & \textbf{Meteor} & \textbf{Div-1} & \textbf{Div-2} & \textbf{Entropy} & \textbf{U} & \textbf{S} & \textbf{L\textsubscript{S}} & \textbf{USL\textsubscript{S}-H}\\ \hline
\multirow{7}{*}{\rotatebox{90}{T5\textsubscript{BoK}}}
& 0.05 &  \underline{51.61} & 29.43 & 18.69 & 12.63 & 4.03 & 4.29 & 16.47 & 0.044 & 0.224 & 9.74 & 0.97 & 0.89 & 0.19 & 0.6748\\ 
& 0.10 &  \textbf{51.74} & \textbf{29.74} & \textbf{19.19} & \textbf{13.24} & \textbf{4.09} & \textbf{4.37} & \underline{16.62} & 0.045 & 0.233 & 9.84 & 0.97 & 0.90 & 0.20 & 0.6793 \\ 
& 0.20  &  51.53 & \underline{29.58} & \underline{19.06} & \underline{13.07} & \underline{4.06} & \underline{4.34} & \textbf{16.71} & 0.046 & 0.231 & \underline{9.85} & 0.97 & 0.90 & 0.21 & \underline{0.6802}\\ 
& 0.30  &  51.08 & 28.91 & 18.44 & 12.55 & 4.00 & 4.26 & 16.58 & 0.046 & \textbf{0.234} & \textbf{9.88} & 0.97 & 0.90 & 0.21 & \textbf{0.6820}\\
& 0.40 &  50.45 & 28.21 & 17.59 & 11.64 & 3.93 & 4.16 & 16.04 & \underline{0.046} & 0.233 & 9.82 & 0.97 & 0.89 & 0.21 & 0.6787\\
& 0.50 &  50.59 & 28.16 & 17.54 & 11.55 & 3.92 & 4.15 & 16.02 & 0.046 & 0.233 & 9.82 & 0.97 & 0.89 & 0.21 & 0.6779\\
& 0.60 &  50.33 & 27.93 & 17.30 & 11.28 & 3.89 & 4.12 & 15.88 & \textbf{0.047} & \underline{0.234} & 9.81 & 0.97 & 0.89 & 0.21 & 0.6764\\

\hline
\multirow{7}{*}{\rotatebox{90}{DialoGPT\textsubscript{BoK}}}
& 0.05  & \underline{49.59} & 27.79 & 17.51 & 11.72 & 3.79 & 4.02 & 16.84 & 0.037 & 0.191 & 9.61 & 0.97 & 0.89 & 0.20 & 0.6765 \\ 
& 0.10 & \textbf{49.62} & 27.91 & 17.68 & 11.90 & 3.81 & 4.05 & 16.84 & 0.038 & 0.195 & 9.65 & 0.97 & 0.89 & 0.21 & 0.6788\\  
& 0.20  & 49.36 & 27.59 & 17.39 & 11.64 & 3.77 & 4.01 & 16.75 & 0.037 & 0.192 & 9.64 & 0.97 & 0.89 & 0.20 & 0.6770 \\ 
& 0.30  & 49.16 & \textbf{29.10} & \textbf{20.00} & \textbf{14.92} & \textbf{4.01} & \textbf{4.35} & \textbf{17.72} & \underline{0.048} & \textbf{0.257} & \textbf{10.19} & 0.97 & 0.89 & 0.31 & \textbf{0.7064} \\ 
& 0.40  & 49.18 & \underline{28.84} & \underline{19.50} & \underline{14.31} & \underline{3.98} & \underline{4.29} & \underline{17.51} & 0.048 & \underline{0.254} & \underline{10.17} & 0.97 & 0.89 & 0.30 & \underline{0.7048}\\ 
& 0.50  & 48.83 & 28.40 & 19.07 & 13.92 & 3.92 & 4.23 & 17.11 & \textbf{0.048} & 0.253 & 10.16 & 0.97 & 0.89 & 0.30 & 0.7048\\ 
& 0.60  & 48.72 & 28.21 & 18.82 & 13.60 & 3.89 & 4.19 & 17.14 & 0.048 & 0.252 & 10.15 & 0.97 & 0.89 & 0.29 & 0.7032\\ 
\hline
\end{tabular}
\caption{Effect of varying $\lambda$ on DailyDialog test performance with $|K_t|=8$.
}
\label{tbl:ablation1}
\end{scriptsize}
\end{table*} 

We observe that for both DailyDialog and Persona-Chat, BoK performs better than BoW in most of the cases. For DailyDialog, DialoGPT\textsubscript{BoK} outperforms DialoGPT\textsubscript{BoW} significantly, which correlates with the automated result shown in Table~\ref{table:result1}. For Persona-Chat, as the generation is conditioned mainly on the persona profiles, the responses are very similar for both models, resulting in a lot of ties. We also observe that BoK loss results in better informativeness, which correlates with the improved specificity (in USL\textsubscript{S}-H) shown in Table~\ref{table:result1} and Table~\ref{tbl:persona}.

\begin{table*}[t]
\begin{scriptsize}
\centering
\begin{tabular}{p{0.5cm}|N|rrrr|rr|r|rr|r|NNNr}
\hline \textbf{Model} & \textbf{$|K_t|$} & \textbf{Bleu-1} & \textbf{Bleu-2} & \textbf{Bleu-3} & \textbf{Bleu-4} & \textbf{Nist-2} & \textbf{Nist-4} & \textbf{Meteor} & \textbf{Div-1} & \textbf{Div-2} & \textbf{Entropy} & \textbf{U} & \textbf{S} & \textbf{L\textsubscript{S}} & \textbf{USL\textsubscript{S}-H}\\ \hline
\multirow{4}{*}{\rotatebox{90}{T5\textsubscript{BoK}}}
& 4 &  \textbf{51.87} & \underline{29.69} & \underline{19.08} & \underline{13.06} & \underline{4.07} & \underline{4.35} & 16.58 & 0.046 & 0.232 & 9.83 & 0.97 & 0.89 & 0.20 & 0.6772 \\ 
& 8 &  \underline{51.74} & \textbf{29.74} & \textbf{19.19} & \textbf{13.24} & \textbf{4.09} & \textbf{4.37} & \underline{16.62} & 0.045 & \underline{0.233} & \underline{9.84} & 0.97 & 0.90 & 0.20 & \textbf{0.6793} \\ 
& 16  &  51.59 & 29.57 & 19.00 & 13.06 & 4.06 & 4.33 & 16.60 & \textbf{0.046} & 0.233 & 9.83 & 0.97 & 0.89 & 0.20 & 0.6780 \\ 
& 24  &  51.66 & 29.58 & 18.96 & 12.96 & 4.06 & 4.33 & \textbf{16.63} & \underline{0.046} & \textbf{0.234} & \textbf{9.85} & 0.97 & 0.89 & 0.20 & \underline{0.6781} \\ 
\hline
\multirow{5}{*}{\rotatebox{90}{DialoGPT\textsubscript{BoK}}}
& 4 &  49.08 & 29.02 & 19.88 & 14.81 & 4.00 & 4.33 & \underline{17.69} & 0.048 & 0.255 & 10.18 & 0.97 & 0.89 & 0.31 & 0.7051 \\  
& 8 &  49.16 & \textbf{29.10} & \textbf{20.00} & \textbf{14.92} & \underline{4.01} & \textbf{4.35} & \textbf{17.72} & 0.048 & \underline{0.257} & \underline{10.19} & 0.97 & 0.89 & 0.31 & \textbf{0.7064} \\ 
& 16  & \underline{49.18} & \underline{29.05} & \underline{19.98} & \underline{14.92} & \textbf{4.01} & \underline{4.35} & 17.62 & 0.048 & \textbf{0.258} & \textbf{10.19} & 0.97 & 0.89 & 0.31 & \underline{0.7054} \\  
& 24  &  \textbf{49.34} & 29.02 & 19.83 & 14.74 & 4.00 & 4.34 & 17.67 & 0.048 & 0.255 & 10.17 & 0.97 & 0.89 & 0.30 & 0.7040 \\
& & & & & & & & & & & & & &\\ 
\hline
\end{tabular}
\caption{Effect of varying maximum number of keyword tokens ($|K_t|$) on DailyDialog test performance. 
}
\label{tbl:ablation2}
\end{scriptsize}
\end{table*} 

\subsection{Human Evaluation}
Table~\ref{tbl:human_eval} shows the human evaluation to compare the impact of BoK and BoW loss on the backbone models. We randomly picked 50 test instances from both DailyDialog and Persona-Chat datasets. Four human evaluators (graduate students proficient in English) were presented with the generated responses from two models (A and B) who reported their judgment (A wins, B wins, or a tie) on various aspects. We asked the evaluators to evaluate five aspects, described as follows. 
\begin{itemize}
    \item \textit{Coherence}: Captures which model produces more contextually coherent responses.
    \item \textit{Engagingness}: Identifies which model generates more engaging or interesting responses.
    \item \textit{Informativeness}: Determines which response contains more knowledge or specific information.
    \item \textit{Interactiveness}: Captures which model produces more interactive responses that encourage the user to continue the conversation.
    \item \textit{Overall}: This is the overall judgment or impression of the evaluator on the given responses.
\end{itemize}
The inter-annotator agreement (Fleiss' kappa) for the overall judgment was 0.81. The Fleiss' kappa for Coherence, Engagingness, Informativeness, and Interactiveness were 0.75, 0.64, 0.63, and 0.60, respectively.



\subsection{Ablation Study}
\label{sec:ablation}
This section analyzes the impact of varying $\lambda$ and $|K_t|$ in the BoK-LM loss. We conduct this ablation study on DailyDialog test data to perform both referenced and reference-free evaluations.

Table~\ref{tbl:ablation1} shows the results of changing $\lambda$ with $|K_t| = 8$ fixed. A higher value of $\lambda$ denotes higher weightage to BoK loss in Eqn.~\ref{eqn:boklm}. For Bleu, Nist, Meteor, Entropy, and USL\textsubscript{S}-H, we observe that increasing $\lambda$ improves the performance up to a certain threshold and then starts declining. In general, T5\textsubscript{BoK} and DialoGPT\textsubscript{BoK} perform well with $\lambda$ values of 0.1 and 0.3, respectively. Div-1 metric measures diversity by counting distinct unigrams. This is why it shows better performance with higher $\lambda$ values, where the context vector $\phi_t$ is learned to predict the keywords with more precision.

Table~\ref{tbl:ablation2} shows the effect of varying the maximum number of keyword tokens ($|K_t|$) in Eqn.~\ref{eqn:bok}, keeping $\lambda$ fixed at 0.1 and 0.3 for T5\textsubscript{BoK} and DialoGPT\textsubscript{BoK}, respectively. Increasing $|K_t|$ makes BoK loss behave more like BoW loss. As a result, we observe DialoGPT\textsubscript{BoK} with $|K_t|=24$ achieves the best Bleu-1 score, which aligns with our expectation. However, this is not the case for T5\textsubscript{BoK}, which we believe happens due to the lower value of $\lambda$ (0.1). Overall, T5\textsubscript{BoK} and DialoGPT\textsubscript{BoK} shows better performance with $|K_t|=8$. 

\begin{table*}[ht]
\centering
\begin{small}
\begin{tabular}{l|G||G|G}
\hline
Turn & Utterance (ground-truth) & Top-8 predicted tokens & Model Generated Response\\ \hline
1 & i need to buy some flowers for my wife . & \textbf{kind}, Ġfresh, Ġred, Ġroses, ros, es, \textbf{Ġkind}, l & we have lots. what \underline{kind} of flowers do you want? \\ \hline
2 & perhaps you 'd be interested in red roses . & \textbf{ros}, \textbf{Ġred}, \textbf{es}, Ġsounds, good, <nok>, expensive, ide & no, i don't like \underline{red} \underline{roses} at all. \\ \hline
3 & what would the roses cost me ? & \textbf{dozen}, Ġprice, \textbf{Ġdollars}, ars, oll, es, ten, Ġred & \underline{\$} 20 per \underline{dozen}. that sounds like a good deal to me. \\ \hline
4 & they 're only \$ 20 a dozen . & \textbf{Ġsounds}, dozen, bad, expensive, \textbf{Ġgood}, price, \textbf{deal}, reasonable & that \underline{sounds} like a \underline{good} \underline{deal}. i'll take a dozen. \\ \hline

\end{tabular}
\caption{ A comprehensive example of dialogue generation with DialoGPT\textsubscript{BoK} on a DailyDialog test instance.}
\label{tbl:ex_dailydialog}
\end{small}
\end{table*}

\begin{table*}[ht]
\centering
\begin{small}
\begin{tabular}{l|G||c|G|G}
\hline
\multicolumn{5}{p{15cm}}{\textbf{Persona of Speaker 1 (P1)}: i am engaged to be married. i am a college graduate. my favorite sport is soccer. i used to play sports in high school.} \\
\multicolumn{5}{p{15cm}}{\textbf{Persona of Speaker 2 (P2)}: i work in a warehouse driving a forklift. i lift weights every weekend. i am big and tall. i like watching horror movies. i like making barbecue.} \\
\hline
\hline
Turn & Utterance (ground-truth) & Persona & Top-8 predicted tokens & Model Generated Response\\ \hline
1 & janice here . what are you up to tonight ? & P2 & \textbf{watching}, \textbf{Ġwatching}, Ġwork, <nok>, work, Ġworking, Ġlove, \textbf{Ġwatch} & hi! i am \underline{watching} a horror movie. how are you? \\ \hline
2 & they call me big shirley . i love scary movies . you ? & P1 & \textbf{Ġlove}, Ġsports, \textbf{Ġwatch}, Ġplay, \textbf{Ġmovies}, \textbf{ovies}, Ġschool, \textbf{Ġwatching} & scary \underline{movies} are great. i \underline{love} \underline{watching} soccer. \\ \hline
3 & well , i am more into sports than i am horror flicks . & P2 & Ġwork, Ġwatch, \textbf{Ġlove}, work, s, Ġworking, Ġsports, Ġgym & i \underline{love} to lift weights. do you lift weights? \\ \hline
4 & i see . i workout lifting weights . you ? & P1 & Ġsports, \textbf{school}, \textbf{Ġplay}, Ġlove, Ġcollege, \textbf{Ġsoccer}, \textbf{soc}, \textbf{cer} & i used to \underline{play} \underline{soccer} in high \underline{school}. you? \\ \hline

\end{tabular}
\caption{A comprehensive example of dialogue generation with DialoGPT\textsubscript{BoK} on a Persona-Chat test instance.}
\label{tbl:ex_persona}
\end{small}
\end{table*}

\section{Discussions}
\subsection{Qualitative Analysis and Interpretability}
In this section, we perform a qualitative analysis of the models trained with BoK-LM loss. Table~\ref{tbl:ex_dailydialog} and Table~\ref{tbl:ex_persona} show comprehensive examples of dialogue generation using DialoGPT\textsubscript{BoK} on a DailyDialog and Person-Chat test instances, respectively. For each dialogue turn, we show the model-generated response. Additionally, we also show the top-8 tokens predicted by the feed-forward neural network (for computing BoK loss) given context vector $\phi_t$. In Table~\ref{tbl:ex_persona}, the ``Persona'' column denotes the persona profile used for response generation. Some tokens have a special character ``Ġ'', which can be interpreted as a space. 

In both tables, we observe an overlapping of tokens between the generated response and the predicted tokens. For example, in Turn 4 of Table~\ref{tbl:ex_dailydialog}, the critical words in the generated response (sounds, good, deal) are in the top-8 predictions. Moreover, for the cases with less overlap, the generated response still aligns with the predicted tokens thematically. For instance, in Turn 3 of Table~\ref{tbl:ex_persona}, the concept of the response matches with the predicted token ``gym''. This refers to the effectiveness of BoK-LM loss in learning the context vector $\phi_t$ that guides the model to generate meaningful responses. Furthermore, $\phi_t$ can be interpreted by looking at the predicted tokens. This is how BoK loss enables post-hoc interpretability in the backbone model.  

\subsection{BoK-LM loss as Reference-Free Metric}
In this section, we study the utility of BoK-LM loss as a reference-free metric for open-domain dialogue evaluation. We conduct our evaluation on various benchmark datasets like USR~\cite{usr}, GRADE~\cite{grade}, PredictiveEngage \cite{predictive-engage}, and FED \cite{fed} that contain human judgments for context-response pairs. 
We use DialoGPT\textsubscript{BoW} and DialoGPT\textsubscript{BoK} to compute the BoW-LM and BoK-LM loss, respectively. BoW-LM and BoK-LM losses are based on cross-entropy loss, where a lower score indicates better quality. As a result, they show a negative correlation with the human scores of the benchmark datasets.

In Table~\ref{tbl:eval}, we can observe that BoK-LM achieves comparable performance to the state-of-the-art metrics on the chit-chat datasets (GRADE-Dailydialog, PredictiveEngage, and FED). However, it shows weaker correlations for knowledge-grounded datasets (USR-Persona and Grade-Convai2) but still performs better than the referenced metrics such as BERTScore, BLEURT, and BERT-RUBER. Moreover, BoK-LM performs better than BoW-LM except for GRADE-DailyDialog dataset. Metrics typically exhibit better performance when applied to the dataset on which they were trained \cite{metric-survey}. Since DialoGPT\textsubscript{BoW} is trained on DailyDialog and has more training data bias than DialoGPT\textsubscript{BoK}, BoW-LM shows superior performance on GRADE-DailyDialog. However, it performs poorly on FED, a relatively difficult dataset. Nevertheless, BoK-LM achieves a decent performance on FED compared to the other metrics. This again verifies that BoK loss is more generalizable than BoW loss.

\begin{table*}[t]
\begin{small}
    
\centering
\begin{tabular}{l|cc|cc|cc|cc|cc}
\hline 
\multirow{2}{*}{{\textbf{Metric}}} & \multicolumn{2}{c|}{\textbf{USR-Persona}} & \multicolumn{2}{c|}{\textbf{GRADE-Convai2}} & \multicolumn{2}{c|}{\textbf{GRADE-Dailydialog}} & \multicolumn{2}{c|}{\textbf{PredictiveEngage}} & \multicolumn{2}{c}{\textbf{FED}}\\ \cline{2-11}
\centering 
& \textbf{P} & \textbf{S}  & \textbf{P} & \textbf{S}  & \textbf{P} & \textbf{S} & \textbf{P} & \textbf{S} & \textbf{P} & \textbf{S} \\ \hline
BLEU-4 & 0.135 & 0.090* & 0.003* & 0.128 & 0.075* & 0.184 & - & - & - & -  \\
METEOR & 0.253 & 0.271 & 0.145 & 0.181 & 0.096* & 0.010* & - & - & - & - \\
BERTScore & 0.152 & 0.122* & 0.225 & 0.224 & 0.129 & 0.100* & - & - & - & - \\
BLEURT & 0.065* & 0.054* & 0.125 & 0.120 & 0.176 & 0.133 & - & - & - & - \\
\scriptsize{BERT-RUBER} & 0.266 & 0.248  & 0.309 & 0.314 & 0.134 & 0.128 & - & - & - & - \\
\hline
MAUDE & 0.345 & 0.298 & 0.351 & -0.304 & -0.036* & -0.073* & 0.104 & 0.060* & 0.018* & -0.094* \\
DEB & 0.291 & 0.373 & 0.426 & \textbf{0.504} & \textbf{0.337} & \textbf{0.363} & 0.516 & 0.580 & \textbf{0.230} & \textbf{0.187} \\
GRADE & 0.358 & 0.352 & \textbf{0.566} & \textbf{0.571} & 0.278 & 0.253 & \textbf{0.600} & 0.622 & 0.134 & 0.118  \\
HolisticEval & 0.087* & 0.113* & -0.030* & -0.010* & 0.025* & 0.020* & 0.368 & 0.365 & 0.122 & 0.125 \\
USR & \textbf{0.440} & \textbf{0.418} & \textbf{0.501} & \textbf{0.500} & 0.057* & 0.057* & \textbf{0.582} & \textbf{0.640} & 0.114 & 0.117 \\
USL-H & \textbf{0.495} & \textbf{0.523} & \textbf{0.443} & 0.457 & 0.108* & 0.093* & \textbf{0.688} & \textbf{0.699} & \textbf{0.201} & \textbf{0.189} \\
Dial-M & \textbf{-0.464} & \textbf{-0.486} & -0.310 & -0.312 & -0.111 & -0.120 & -0.570 & -0.592 & -0.127 & -0.097 \\
\hline
BoW-LM & -0.156 & -0.124 & -0.286 & -0.252 & \textbf{-0.419} & \textbf{-0.443} & -0.534 & -0.572 & -0.048* & -0.082*\\
BoK-LM & -0.261 & -0.255 & -0.318 & -0.301 & \textbf{-0.367} & \textbf{-0.383} & -0.581 & \textbf{-0.632} & \textbf{-0.135} & \textbf{-0.151}\\
\hline
\end{tabular}
\caption{Comparison of dialogue evaluation metrics with top-3 scores highlighted in bold. P and S indicate Pearson and Spearman's coefficients, respectively. All values are statistically significant to $p < 0.05$, unless marked by *.}
\label{tbl:eval}
\end{small}
\end{table*}
\section{Conclusion}
This paper proposes Bag-of-Keywords (BoK) loss, a novel auxiliary loss for training open-domain dialogue systems. The main idea of BoK loss is to improve the generalizability of Bag-of-Words (Bow) loss by predicting only the keywords or the core idea of the next response. We show that BoK loss enhances the generative performance of the vanilla T5 and DialoGPT models on the DailyDialog and Persona-Chat datasets when trained with BoK-LM loss. We also notice an improvement in the specificity of the generated response with the inclusion of BoK loss. We discuss the notion of interpretability that comes with the incorporation of BoK loss with comprehensive examples. Finally, we show that BoK-LM loss shows a moderate performance as a reference-free dialogue evaluation metric. In future work, we want to explore better keyword extraction methods and study the applicability of BoK loss in other NLG tasks.

\bibliography{main}

\appendix

\section{Appendix}
\label{sec:appendix}

\subsection{Additional Implementation Details}
\label{sec:app1}
For training data preparation related to BoK loss, we first extract the keywords from the next utterance using YAKE!~\cite{yake2, yake}. It outputs the keywords as a list with a decreasing order of relevance. We concatenate this list of keywords into a string and then tokenize it using the T5/GPT tokenizer. We consider the top-k tokens based on the maximum token limit ($|K_t|$). There are instances where the YAKE! could not find any keywords. In those cases, we add a special token (\texttt{<nok>}) in the label. In other words, the model is trained to predict \texttt{<nok>} for generic responses with no keywords.

As discussed, we studied the effectiveness of our proposed BoK loss by applying it to T5 and DialoGPT. We performed our experiment with DailyDialog and Persona-Chat datasets. For each dataset, we train a separate T5 and DialoGPT model. The two datasets and models only support the English language. The best model was selected for each training based on the validation loss. The training time of all the models is around 12-20 hours. Since we do not have any sampling during training and use a fixed seed (10), the models are reproducible. Furthermore, we generate the responses using beam search with a fixed configuration (described in Section \ref{sec:det}). Because of that, we report the results of the model with a single run since they are deterministic. We use four data-specific baselines - DilaoFlow ($\approx$ 900M parameters)~\footnote[3]{\href{https://github.com/ictnlp/DialoFlow}{github.com/ictnlp/DialoFlow}}, DialogVED ($\approx$ 400M parameters)~\footnote[4]{\href{https://github.com/lemuria-wchen/DialogVED}{github.com/lemuria-wchen/DialogVED}}, DialoGEN ($\approx$ 900M parameters)~\footnote[5]{\href{https://github.com/SuvodipDey/DialoGen}{github.com/SuvodipDey/DialoGen}}, and TransferTransfo ($\approx$ 200M parameters)~\footnote[6]{\href{https://github.com/huggingface/transfer-learning-conv-ai}{github.com/huggingface/transfer-learning-conv-ai}}. Codes of all the baselines are publicly available and have free license.

The referenced evaluation of the generated dialogues was conducted following the evaluation of DSTC7 Task 2~\footnote[7]{\href{https://github.com/mgalley/DSTC7-End-to-End-Conversation-Modeling/tree/master/evaluation/src}{github.com/mgalley/DSTC7-End-to-End-Conversation-Modeling/tree/master/evaluation/src}}. We used two different models to compute the BoK-LM loss in Table~\ref{tbl:eval}. For the knowledge-grounded datasets (USR-Persona, GRADE-Convai2), we used the DialoGPT\textsubscript{BoK} model trained on the Persona-Chat dataset. For the chit-chat datasets (GRADE-DailyDialog, Predictive Engage, and FED), we utilized the DialoGPT\textsubscript{BoK} model trained on the DailyDialog dataset. The same process is followed to compute the BoW-LM loss as well. 



\begin{table*}[t]
\centering
\begin{small}
\begin{tabular}{l|G||G|G}
\hline
Turn & Utterance (ground-truth) & Top-8 predicted tokens (BoK) & Top-8 predicted tokens (BoW)\\ \hline
1 & i need to buy some flowers for my wife . & (kind, 0.1113), (Ġfresh, 0.0913), (Ġred, 0.0629), (Ġroses, 0.0332), (ros, 0.0304), (es, 0.0277), (Ġkind, 0.0249), (l, 0.0199) & (Ġ?, 0.0856), (Ġroses, 0.0649), (Ġyou, 0.0382), (Ġkind, 0.0347), (Ġ., 0.0314), (Ġlike, 0.0295), (how, 0.0234), (Ġare, 0.0224)
 \\ \hline
2 & perhaps you 'd be interested in red roses . & (ros, 0.2161), (Ġred, 0.2063), (es, 0.0894), (Ġsounds, 0.0227), (good, 0.0147), (<nok>, 0.0118), (expensive, 0.0083), (ide, 0.0079)
 & (Ġ?, 0.0896), (Ġ., 0.0816), (Ġ,, 0.0686), (Ġlike, 0.0453), (Ġi, 0.0379), (Ġroses, 0.0285), (Ġthey, 0.0215), (how, 0.0161) \\ \hline
3 & what would the roses cost me ? & (dozen, 0.7592), (Ġprice, 0.0139), (Ġdollars, 0.0111), (ars, 0.009), (oll, 0.006), (es, 0.0037), (ten, 0.0033), (Ġred, 0.0032)
 & (Ġ., 0.121), (Ġeach, 0.1), (Ġdollars, 0.0324), (Ġper, 0.0272), (Ġ\$, 0.0264), (Ġdozen, 0.0259), (they, 0.0228), (the, 0.0223)
 \\ \hline
4 & they 're only \$ 20 a dozen . &  (Ġsounds, 0.1743), (dozen, 0.1095), (bad, 0.0831), (expensive, 0.0669), (Ġgood, 0.0486), (price, 0.0395), (deal, 0.0219), (reasonable, 0.0185)
 & (Ġ., 0.0788), (Ġ?, 0.0409), (Ġ,, 0.0356), (that, 0.0326), (Ġi, 0.0251), (Ġa, 0.0233), (i, 0.0231), (how, 0.0226) \\ \hline

\end{tabular}
\caption{Comparison of predicted tokens on a DailyDialog test instance.}
\label{tbl:ex_dailydialog2}
\end{small}
\end{table*}

\subsection{Related Works on Open-domain Dialogue Evaluation}
\label{sec:rel_eval}
Since we study the usefulness of our proposed loss as a reference-free metric, we add a short literature survey on open-domain dialogue evaluation. There are primarily two kinds of dialogue evaluation metrics- i) referenced and ii) reference-free. In referenced metrics, the generated response is compared with one or more reference utterances to evaluate its goodness. The most popular referenced metrics are word-overlapping based metrics like BLEU \cite{bleu}, NIST \cite{nist}, METEOR \cite{meteor}, Diversity \cite{diversity}, and Entropy \cite{entropy}. There are also learning-based referenced metrics like ADEM \cite{adem}, RUBER \cite{ruber}, BERT-RUBER \cite{bert-ruber}, PONE \cite{pone}, BERTScore \cite{bertscore}, BLEURT \cite{bleurt}, etc. Conversely, the reference-free metrics are designed to evaluate dialogues without any references. As collecting good-quality references is expensive and needs human effort, most of the recent research focuses on developing reference-free metrics. Most of the methods formulate the dialogue evaluation problem as a classification task and use the classification score as the metric \cite{maude, deb, grade, dynaeval}. Metrics such as USR \cite{usr}, USL-H \cite{usl-h}, FED \cite{fed}, HolisticEval \cite{holistic}, D-score \cite{dscore}, and QualityAdapt \cite{qualityadapt} combine various sub-metrics to provide more holistic evaluation. Dial-M \cite{dial-m} adopts a masking-based approach that utilizes masked language modeling (MLM) loss as the evaluation score. Metrics like $IM^2$ \cite{im2} leverage various evaluation metrics to enhance the evaluation of different dialogue aspects.

\subsection{Comparison of top-k Predicted Tokens (BoK vs. BoW)}
Table~\ref{tbl:ex_dailydialog2} shows the top-k tokens associated with the BoW and BoK loss (along with the probability scores) for the examples shown in Table~\ref{tbl:ex_dailydialog}. We use the DialoGPT\textsubscript{BoK} and DialoGPT\textsubscript{BoW} to find the top-k BoK and BoW, respectively. We can observe that the top-8 tokens associated with the BoW loss contain a lot of punctuation and stopwords as they are trained to predict all the words/tokens of the next utterance. In contrast, the top-k tokens associated with the BoK are more aligned with the conversation topic. For example, in Turn 4 of Table~\ref{tbl:ex_dailydialog2}, all the tokens predicted by the BoK method are relevant and can potentially steer the conversation in a meaningful direction. However, for the BoW method, the predicted words are mostly punctuations and stopwords.

\end{document}